\documentclass[a4paper,fleqn]{cas-dc}

\usepackage[numbers,sort&compress]{natbib}
\usepackage[commandnameprefix=always]{changes}
\usepackage{subcaption}

\makeatletter
\renewcommand\printorcid{}
\makeatother

\usepackage{tabularx} 
\usepackage{amsthm}
\usepackage{array}
\usepackage{float}
\usepackage{algorithm}
\usepackage{algpseudocode}
\usepackage{booktabs}
\usepackage{threeparttable}
\usepackage[normalem]{ulem}
\theoremstyle{remark}          
\hypersetup{
  colorlinks=false,        
 pdfborder={0 0 1},       
            citebordercolor={0 1 0}, 
            linkbordercolor={1 0 0}, 
            urlbordercolor={0 1 1}   
}

\begin{document}
\let\WriteBookmarks\relax
\def\floatpagepagefraction{1}
\def\textpagefraction{.001}

\shorttitle{QuadKAN: KAN-Enhanced Quadruped Motion Control via End-to-End Reinforcement Learning}

\shortauthors{Allen Wang and Gavin Tao}

\title [mode = title]{QuadKAN: KAN-Enhanced Quadruped Motion Control via End-to-End Reinforcement Learning}                      


\author[1]{Allen Wang}[style=chinese]
\ead{}
\credit{}

\author[2]{Gavin Tao}[style=chinese]
\cormark[1]
\ead{}
\credit{}


\cortext[cor1]{Corresponding author}

\begin{abstract}
We address vision-guided quadruped motion control with reinforcement learning (RL) and highlight the necessity of combining proprioception with vision for robust control. We propose \emph{QuadKAN}, a spline-parameterized cross-modal policy instantiated with Kolmogorov--Arnold Networks (KANs). The framework incorporates a spline encoder for proprioception and a spline fusion head for proprioception–vision inputs. This structured function class aligns the state-to-action mapping with the piecewise-smooth nature of gait, improving sample efficiency, reducing action jitter and energy consumption, and providing interpretable posture–action sensitivities. We adopt Multi-Modal Delay Randomization (MMDR) and perform end-to-end training with Proximal Policy Optimization (PPO). Evaluations across diverse terrains, including both even and uneven surfaces and scenarios with static or dynamic obstacles, demonstrate that \emph{QuadKAN} achieves consistently higher returns, greater distances, and fewer collisions than state-of-the-art (SOTA) baselines. These results show that spline-parameterized policies offer a simple, effective, and interpretable alternative for robust vision-guided locomotion. A repository is hosted at \url{https://github.com/allen-quad-robot/quadkan}, to be made available upon acceptance.
\end{abstract}


\begin{highlights}
\item \textbf{State-of-the-art QuadKAN.} We propose a structured cross-modal policy for quadrupedal locomotion, parameterized by splines and instantiated via KANs. To the best of our knowledge, this is the first vision-driven, KAN-enhanced DRL framework for quadruped motion control.

\item \textbf{Effective proprioception–vision representation.} We design a compact input representation in which an MLP embeds proprioceptive states and a CNN patchifies depth images into tokens tailored for fusion. This provides egocentric stability and foresightful obstacle awareness while reducing visual sensitivity and enhancing training efficiency.  

\item \textbf{KAN effectiveness.} We introduce KANs into both the proprioceptive encoder and the fusion head. The spline-based inductive bias aligns with the piecewise-smooth structure of gait dynamics, yielding improved sample efficiency, policy stability, and interpretability compared with unstructured regressors. 

\item \textbf{PPO training robustness.} We introduce an MMDR protocol with end-to-end PPO optimization. A compact, state-centric reward balances task-aligned progress, energy efficiency, and safety, enabling stable learning and consistent performance.  

\item \textbf{Comprehensive evaluation.} Extensive experiments across trained, static-rugged, and dynamic-obstacle terrains demonstrate consistent gains over state-of-the-art (SOTA) baselines in return, collision avoidance, and distance traveled, validating the effectiveness of the proposed framework.  
\end{highlights}

\begin{keywords}
Quadrupedal Robots\sep Vision-driven Locomotion \sep Reinforcement Learning\sep Multi-modal Fusion\sep End-to-end Policy Learning
\end{keywords}

\maketitle

\section{Introduction}
\label{sec:intro}
Legged robots offer mobility where wheeled platforms fail, such as stairs, rubble, soft substrates, and cluttered indoor–outdoor settings, enabling applications in inspection, search and rescue, agriculture, and planetary exploration \citep{fan2024review}. Robust locomotion control is therefore a foundational capability for practical quadrupedal systems, underpinning safe navigation and dependable operation across diverse terrains and disturbances \citep{carpentier2021recent}. Deep reinforcement learning (DRL) has emerged as a compelling paradigm for such control because it optimizes closed-loop policies through interaction and can produce adaptive behaviors \citep{zhang2022deepreinforcementlearningforreal}. 

A substantial body of prior work has focused on training \emph{blind} controllers that rely exclusively on proprioceptive inputs such as inertial measurement units (IMUs) and joint feedback \citep{li2021survey}. While these blind policies can traverse uneven and unknown terrains through large-scale simulation and domain randomization, they inherently lack foresight: without exteroceptive input, they respond only upon contact and struggle to proactively avoid obstacles or plan foot placement on irregular ground. Vision complements proprioception by providing anticipatory geometric information, enabling early detection of distant obstacles and terrain changes \citep{wang2025locomamba}. As a result, cross-modal policies that integrate proprioception with depth imaging have gained prominence, facilitating safer and more efficient locomotion through earlier trajectory adjustments.

Most existing cross-modal pipelines adopt multilayer perceptrons (MLPs) for the proprioceptive encoder and for the decision head that fuses proprioception with vision. While effective, this unstructured regressor has well-known drawbacks in the locomotion setting \citep{imai2022vision}: it provides limited inductive bias for the piecewise-smooth nature of gait dynamics (smooth within a phase, sharp transitions at contact events), couples inputs globally rather than locally, and offers little interpretability about how posture variables influence joint commands. These properties can manifest as either over-smoothing or oscillatory actions near phase boundaries, increased sensitivity to asynchronous sensing, and difficulty diagnosing failure cases. Moreover, improving expressivity typically requires widening or deepening the MLP, which can inflate parameters and on-board compute without addressing the underlying mismatch in function class \citep{singh2022reinforcement}.

This paper introduces \emph{QuadKAN}, a vision-guided end-to-end deep reinforcement learning (DRL) framework that replaces unstructured regressors with a spline-parameterized policy class instantiated via Kolmogorov--Arnold Networks (KANs) \citep{liu2024kan}. KAN layers implement mappings using spline bases with local support, naturally capturing the piecewise-smooth structure of gait. We employ KANs in two components: (i) a proprioceptive spline encoder that maps IMU and joint features to compact tokens aligned with intra-phase smoothness and inter-phase transitions, and (ii) a spline fusion head that integrates proprioceptive tokens with depth features while maintaining a lightweight, attention-free compute profile. This structured policy class promotes stable, locally controlled responses, reduces action jitter, and enables actionable interpretability through coefficient analysis and local-sensitivity visualizations \citep{somvanshi2024survey}.

A practical challenge in vision-guided control is asynchrony: sensor, inference, and actuation latencies differ across modalities—particularly for visual processing—resulting in misalignment between proprioceptive and visual streams. To mitigate this, Multi-Modal Delay Randomization (MMDR) is adopted, which randomizes the temporal selection of both proprioceptive and visual inputs during training to emulate real-world, modality-dependent latencies. Policies are trained end-to-end using Proximal Policy Optimization (PPO) under the same MMDR protocol and domain randomization settings as prior work, enabling fair comparisons \citep{imai2022vision}.

\textbf{The primary contributions of this work are as follows:}
\begin{enumerate}[1)]
\item \textbf{State-of-the-art QuadKAN.} We propose a structured cross-modal policy for quadrupedal locomotion, parameterized by splines and instantiated via KANs. To the best of our knowledge, this is the first vision-driven, KAN-enhanced DRL framework for quadruped motion control.

\item \textbf{Effective proprioception–vision representation.} We design a compact input representation in which an MLP embeds proprioceptive states and a CNN patchifies depth images into tokens tailored for fusion. This provides egocentric stability and foresightful obstacle awareness while reducing visual sensitivity and enhancing training efficiency.  

\item \textbf{KAN effectiveness.} We introduce KANs into both the proprioceptive encoder and the fusion head. The spline-based inductive bias aligns with the piecewise-smooth structure of gait dynamics, yielding improved sample efficiency, policy stability, and interpretability compared with unstructured regressors. 

\item \textbf{PPO training robustness.} We introduce an MMDR protocol with end-to-end PPO optimization. A compact, state-centric reward balances task-aligned progress, energy efficiency, and safety, enabling stable learning and consistent performance.  

\item \textbf{Comprehensive evaluation.} Extensive experiments across trained, static-rugged, and dynamic-obstacle terrains demonstrate consistent gains over state-of-the-art (SOTA) baselines in return, collision avoidance, and distance traveled, validating the effectiveness of the proposed framework.  
\end{enumerate}

Section~\ref{sec:related} reviews prior work on blind and cross-modal locomotion. Section~\ref{sec:method} details the QuadKAN framework, including the problem formulation, proprioceptive spline encoder, KAN-based fusion mechanism, and the MMDR training strategy. Section~\ref{sec:implementation} describes the system architecture, network configurations, and training pipeline. Section~\ref{sec:evaluation} presents quantitative and qualitative evaluations across diverse terrains and perturbations, along with ablation studies, interpretability analyses, and efficiency profiling. Finally, Section~\ref{sec:conclusion} concludes the paper and outlines future research directions.

\section{Related Work}
\label{sec:related}

\subsection{Learning-Based Quadrupedal Locomotion}
Quadrupedal locomotion has been approached through both model-based and learning-based methods. Model-based pipelines typically rely on template dynamics, trajectory optimization over centroidal or full-body models, and model predictive control with explicit contact handling~\citep{miura1984dynamic,bledt2018cheetah,grandia2019feedback,di2018dynamic,ding2019real,carius2019trajectory}. While these methods can produce precise behaviors, they require accurate modeling and extensive parameter tuning, which limits scalability in complex or unstructured environments. In contrast, model-free RL optimizes closed-loop policies directly from interaction, often combined with domain randomization, enabling robust gait generation over uneven terrain and under external disturbances~\citep{tan2018sim,hwangbo2019learning,kumar2021rma}. However, much of this progress has focused on \emph{proprioception-only} controllers that lack exteroceptive input, constraining their ability to anticipate obstacles and perform precise foothold planning~\citep{xie2022glide}.

\subsection{Vision-Driven RL for Quadrupedal Locomotion}
Vision provides look-ahead geometric information that complements proprioception, enabling early perception of distant obstacles and anticipation of terrain changes prior to contact. Recent work integrates depth or image observations into RL pipelines, either end-to-end or through hierarchical structures~\citep{jain2019hierarchical,yu2021visual,duan2024learning,fahmi2022vital,han2025multimodal}. While effective, commonly used fusion backbones present notable trade-offs: simple concatenation underutilizes spatial structure and lacks inductive bias; recurrent models suffer from vanishing gradients and limited capacity for long-horizon dependencies; and Transformer-based policies, though expressive, incur quadratic memory and computational costs with respect to token count, thereby constraining sequence length, spatial resolution, and training efficiency~\citep{yang2021learning}. These limitations motivate the design of lightweight, structured policy architectures that capture locomotion-specific regularities without incurring the overhead of attention mechanisms.

\subsection{Policy Function Classes and Structured Priors}
Most cross-modal locomotion pipelines encode proprioceptive input using MLPs and implement fusion or decision heads with additional MLP layers. Although widely adopted, these unstructured regressors offer limited inductive bias for the piecewise-smooth nature of gait, which is typically smooth within a phase and exhibits sharp transitions at contact events \citep{imai2022vision}. Their global coupling of inputs can hinder local control, reduce interpretability of posture--action relationships, and result in either over-smoothed outputs or oscillatory actions near phase boundaries. Increasing model width or depth inflates parameter count and on-board computational cost without resolving the underlying function-class mismatch \citep{yang2021learning}. Spline-parameterized networks provide a structured alternative by composing basis functions with local support, thereby enabling explicit control over smoothness and curvature, and naturally aligning with phase-wise dynamics. KANs instantiate this principle by parameterizing layers with spline bases, yielding localized responses and interpretable sensitivity profiles~\citep{liu2024kan}. To the best of our knowledge, this is the first work to explore quadrupedal locomotion using KANs for both the proprioceptive encoder and the proprioception-vision fusion head.

\subsection{Asynchrony, Delay, and Domain Randomization}
Latency varies across sensing, inference, and actuation, with visual processing introducing particularly significant delays. This results in asynchronous multi-modal inputs and misalignment between proprioception and exteroception in real-world systems. Domain randomization is widely used to bridge the sim-to-real gap by perturbing dynamics, contacts, and visual attributes~\citep{imai2022vision}; however, modality-dependent delays require dedicated treatment. MMDR addresses this by randomizing the temporal selection of proprioceptive and visual streams during training, emulating real-world asynchrony and improving robustness to cross-modal misalignment~\citep{yang2021learning}. Complementary work has explored continuous-time RL and delayed Markov decision processes (MDPs) for constant-latency scenarios~\citep{wang2021survey,singh2022reinforcement}. In our framework, MMDR is integrated into an end-to-end PPO pipeline to account for modality-dependent delays while maintaining a lightweight, attention-free fusion mechanism.

\section{Methodology}
\label{sec:method}

\begin{figure*}[t]
  \centering
  \includegraphics[width=1.0\linewidth]{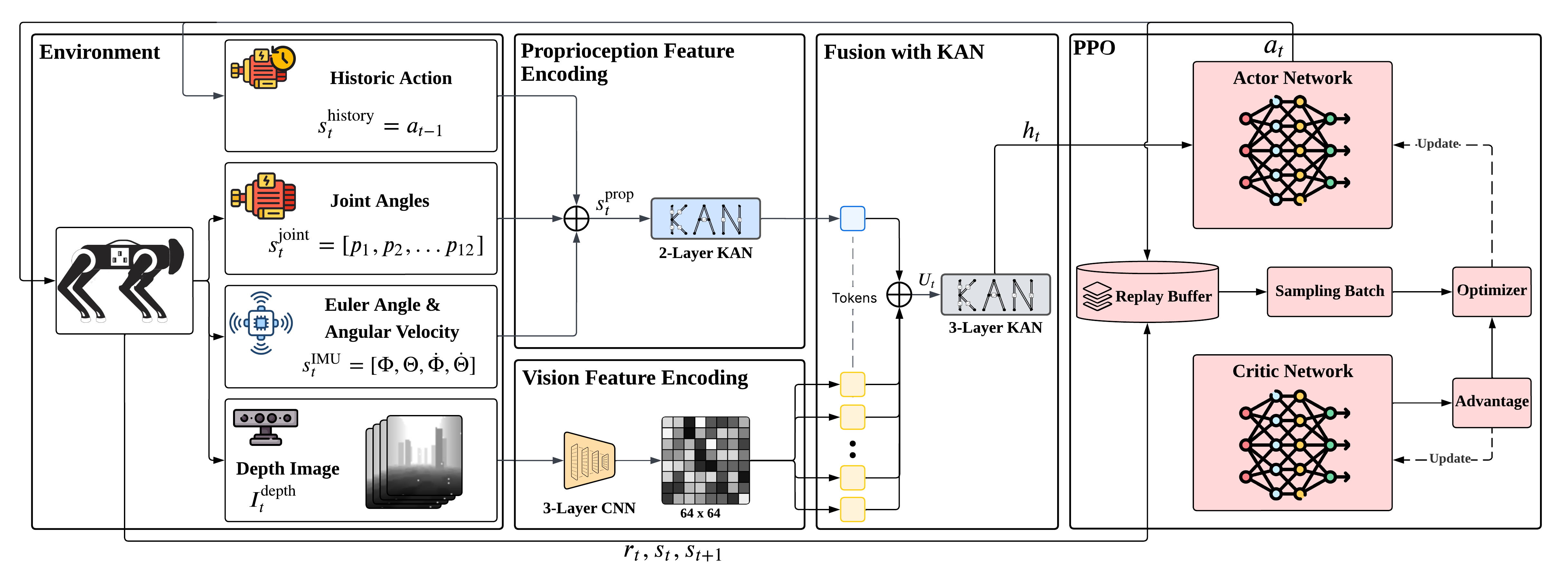}
  \caption{Overall architecture of \emph{QuadKAN}. Proprioceptive states are mapped by a KAN into a compact token; a lightweight CNN patchifies depth into spatial tokens. Tokens are projected to a common width and fused by a KAN-based spline head without attention. Policy and value heads are trained end-to-end with PPO.}
  \label{fig:quadkan-arch}
\end{figure*}

Figure~\ref{fig:quadkan-arch} summarizes \emph{QuadKAN}. The design follows a standard end-to-end DRL pipeline but replaces unstructured regressors with a spline-parameterized policy class instantiated via KAN. The policy consumes proprioception and depth inputs, performs attention-free fusion, and is optimized using PPO under the MMDR protocol.

\subsection{Cross-Modal Observations and Tokenization}
At time $t$, the agent receives a composite observation
\begin{equation}
o_t \;=\; \big\{\, s^{\mathrm{prop}}_t,\ I^{\mathrm{depth}}_t \,\big\},
\label{eq:obs}
\end{equation}
where $s^{\mathrm{prop}}_t$ denotes proprioceptive feedback (IMU, joint states, and optionally the previous action) and $I^{\mathrm{depth}}_t$ is a first-person depth image.

Modality-specific encoders map each input to a latent token space:
\begin{align}
z^{\mathrm{prop}}_t &= f_{\mathrm{KAN}}\!\left(s^{\mathrm{prop}}_t\right) \in \mathbb{R}^{d_p}, 
\label{eq:prop-encoding}\\
z^{\mathrm{vis}}_t  &= f_{\mathrm{CNN}}\!\left(I^{\mathrm{depth}}_t\right) \in \mathbb{R}^{N \times d_v},
\label{eq:vis-encoding}
\end{align}
where $z^{\mathrm{vis}}_t$ is a sequence of $N$ visual tokens obtained by patchifying the image into $P\times P$ blocks (assuming $P \mid H$ and $P \mid W$), so that
\[
N \;=\; \Big(\tfrac{H}{P}\Big)\!\cdot\!\Big(\tfrac{W}{P}\Big).
\]

Both modalities are projected to a common width $d$ and concatenated along the token dimension to form a unified cross-modal stream:
\begin{align}
\tilde z^{\mathrm{prop}}_t &= W_p\, z^{\mathrm{prop}}_t \in \mathbb{R}^{d},  \\
\tilde z^{\mathrm{vis}}_t  &= z^{\mathrm{vis}}_t\, W_v \in \mathbb{R}^{N \times d},  \\
U_t &= \big[\, \tilde z^{\mathrm{prop}}_t \ ;\ \tilde z^{\mathrm{vis}}_t \,\big] \in \mathbb{R}^{(1+N)\times d}.
\end{align}

We augment $U_t$ with learned spatial/positional embeddings $E_{\mathrm{pos}}^{\mathrm{spat}} \in \mathbb{R}^{(1+N)\times d}$ (a reserved index for the proprioceptive token and a 2D grid for the $N$ visual tokens) and modality tags $E_{\mathrm{mod}} \in \mathbb{R}^{(1+N)\times d}$ (distinct tags for proprioception vs.\ vision). The input to the backbone is
\begin{equation}
\hat U_t \;=\; \mathrm{LN}\!\left( U_t \;+\; E_{\mathrm{pos}}^{\mathrm{spat}} \;+\; E_{\mathrm{mod}} \right),
\end{equation}
where $\mathrm{LN}$ denotes layer normalization applied across the embedding dimension.

\subsection{Proprioceptive Spline–KAN Encoder}
\begin{figure}[t]
  \centering
  \includegraphics[width=\linewidth]{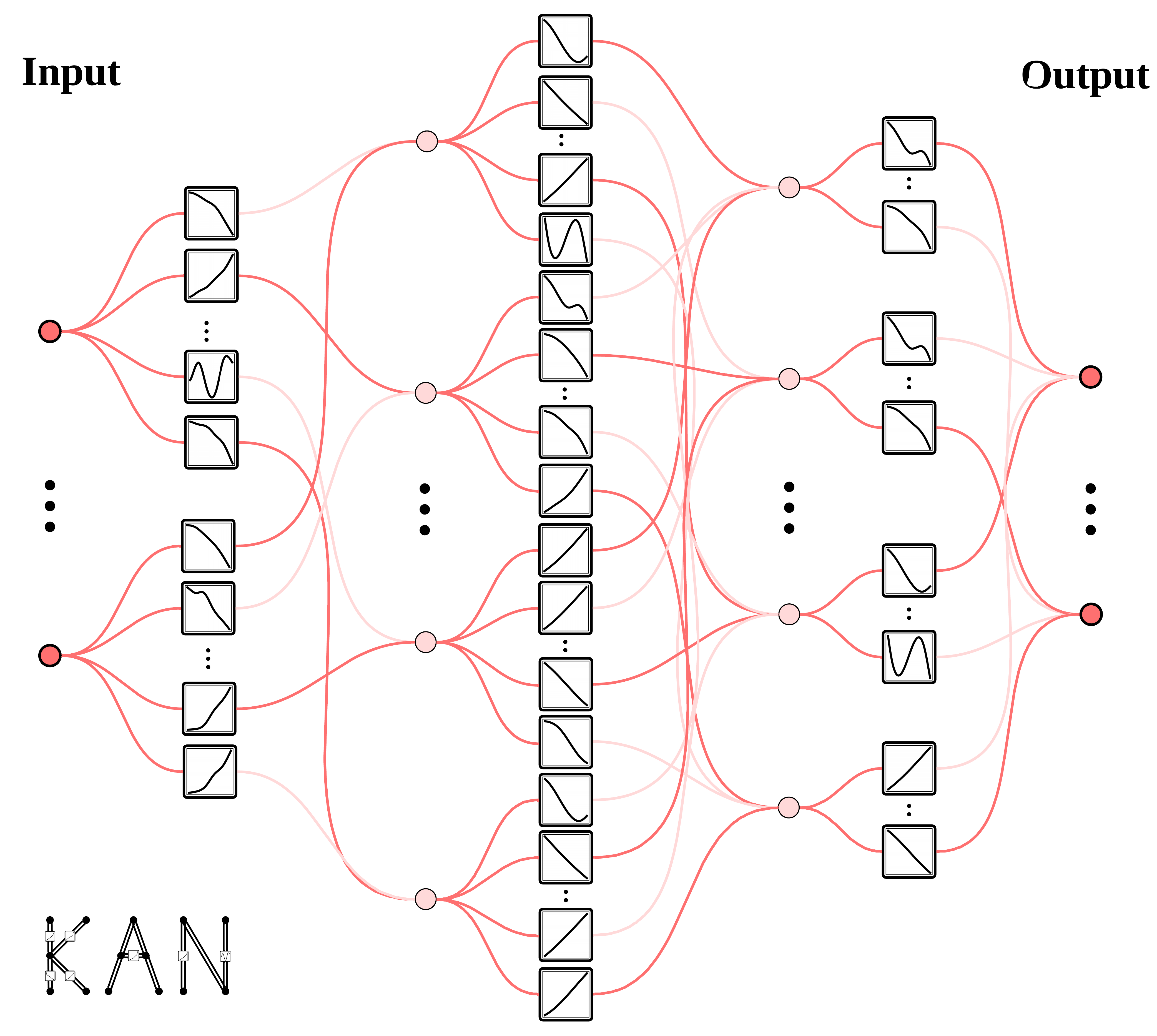}
  \caption{Spline–KAN encoder. Each unit projects the input to a scalar, expands it over a compact bank of B-spline basis functions, aggregates with learnable coefficients, and applies a smooth nonlinearity. Stacking units produces tokens with localized, phase-aware sensitivities. Optional curvature and Jacobian regularizers promote smooth, low-jitter actuation.}
  \label{fig:kan-arch}
\end{figure}

Gait control is piecewise smooth: trajectories are locally smooth within contact phases and exhibit sharp transitions at foot-strike/lift-off. We encode this inductive bias by mapping proprioception using spline bases with local support. The KAN architecture is shown in Fig.~\ref{fig:kan-arch}. Each KAN unit combines a learnable scalar projection with a B-spline expansion and a pointwise nonlinearity, yielding localized and interpretable responses aligned with phase-wise dynamics.

Let $x\in\mathbb{R}^{d_{\mathrm{in}}}$ denote the proprioceptive vector. For unit $j$,
\begin{equation}
\phi_j(x)
= \sigma\!\Big(b_j + \sum_{m=1}^{M} w_{j m}\, B_m\!\big(a_j^\top x + c_j\big)\Big),
\qquad j=1,\dots,d_p,
\label{eq:kan-unit}
\end{equation}
where $\{B_m\}_{m=1}^M$ are fixed B-spline basis functions (degree $q$, knot vector $\mathcal{K}$), $a_j\!\in\!\mathbb{R}^{d_{\mathrm{in}}}$ and $c_j\!\in\!\mathbb{R}$ define the scalar input to the basis, $w_{jm}\!\in\!\mathbb{R}$ are learnable coefficients, $b_j\!\in\!\mathbb{R}$ is a bias, and $\sigma(\cdot)$ is a smooth activation. Stacking such layers yields
$
f_{\mathrm{KAN}}:\mathbb{R}^{d_{\mathrm{in}}}\!\to\!\mathbb{R}^{d_p}
$
with localized responses.

To promote stable control and suppress high-frequency actuation, we regularize the spline coefficients via a discrete curvature penalty and control the sensitivity via a Jacobian (Lipschitz-style) term:
\begin{subequations}
\label{eq:kan-reg}
\begin{align}
\mathcal{R}_{\mathrm{curv}}
&= \lambda_c \sum_{j=1}^{d_p}\sum_{m=2}^{M-1}
\big(w_{j,m+1}-2w_{j,m}+w_{j,m-1}\big)^2,\\
\mathcal{R}_{\mathrm{jac}}
&= \lambda_L\, \mathbb{E}_{x\sim\mathcal{D}}\!\Big[\big\|\tfrac{\partial \phi(x)}{\partial x}\big\|_F^2\Big],\\
\mathcal{R}_{\mathrm{spline}} &= \mathcal{R}_{\mathrm{curv}}+\mathcal{R}_{\mathrm{jac}},
\end{align}
\end{subequations}
where $\phi(x)=[\phi_1(x),\dots,\phi_{d_p}(x)]^\top$ and $\mathcal{D}$ is the training distribution.

\subsection{SplineFusion Decision KAN Head}
Cross-modal fusion is implemented with a lightweight, token-wise KAN head that maps the concatenated stream $\hat U_t\in\mathbb{R}^{(1+N)\times d}$ to fused features without attention, preserving linear complexity in the token count and the spline prior:
\begin{equation}
Y_t \;=\; G_{\mathrm{KAN}}(\hat U_t) \in \mathbb{R}^{(1+N)\times d}.
\end{equation}
Let the first token correspond to proprioception and the remaining $N$ tokens to vision. We form a compact summary by averaging visual tokens and concatenating with the proprioceptive token:
\begin{equation}
\bar y^{\mathrm{vis}}_t
= \frac{1}{N}\sum_{i=1}^{N} Y^{\mathrm{vis}}_{t,i},
\qquad
h_t
= f_{\mathrm{head}}\!\Big(\big[Y^{\mathrm{prop}}_{t,1}\ ;\ \bar y^{\mathrm{vis}}_t\big]\Big)
\in \mathbb{R}^{d_h},
\end{equation}
where $f_{\mathrm{head}}$ is a compact projection (KAN block). The representation $h_t$ feeds the policy and value heads in the RL module.

\subsection{Policy Optimization}
We cast control as an MDP and optimize an end-to-end PPO agent \citep{schulman2017proximal}. The policy is a Tanh-squashed Gaussian and the critic is a scalar value function:
\begin{align}
\tilde a_t &\sim \mathcal{N}\!\big(\mu_\theta(h_t),\,\mathrm{diag}(\sigma^2_\theta(h_t))\big), \\
a_t &= a_{\max}\,\tanh(\tilde a_t), \\
V_\phi(h_t) &\approx \mathbb{E}\!\Big[\sum_{l\ge 0}\gamma^{\,l}\, r_{t+l}\,\Big|\,h_t\Big].
\end{align}
We compute advantages with GAE($\lambda$):
\begin{align}
\delta_t &= r_t + \gamma V_\phi(h_{t+1}) - V_\phi(h_t),\\
A_t &= \sum_{l=0}^{T-1-t} (\gamma\lambda)^l\, \delta_{t+l}, \qquad
\hat R_t \;=\; A_t + V_\phi(h_t).
\end{align}
Let $\rho_t(\theta) = \frac{\pi_\theta(a_t\mid h_t)}{\pi_{\theta_{\mathrm{old}}}(a_t\mid h_t)}$ denote the importance ratio, where $\pi_\theta(a_t\mid h_t)$ is computed with the change-of-variables correction for Tanh-squashing.\footnote{If $a_t=a_{\max}\tanh(\tilde a_t)$ with $\tilde a_t \sim \mathcal{N}(\mu_\theta,\Sigma_\theta)$, then
$\log \pi_\theta(a_t\!\mid\!h_t)
= \log \mathcal{N}(\tilde a_t;\mu_\theta,\Sigma_\theta)
- \sum_i \log\!\big(a_{\max,i}(1-\tanh^2(\tilde a_{t,i}))\big)$.}
The clipped surrogate and auxiliary terms are
\begin{align}
\mathcal{L}_{\pi}(\theta)
&= -\,\mathbb{E}\bigg[
\min\Big(\rho_t A_t,\;
\mathrm{clip}(\rho_t,1-\epsilon,1+\epsilon)\,A_t\Big)
\bigg],\\
\mathcal{L}_{V}(\phi)
&= \mathbb{E}\!\big[(V_\phi(h_t)-\hat R_t)^2\big],\\
\mathcal{L}_{H}(\theta)
&= -\,\mathbb{E}\!\big[\mathcal{H}(\pi_\theta(\cdot\mid h_t))\big].
\end{align}

The total loss minimized during training combines policy, value, entropy, and spline regularization:
\begin{equation}
\label{eq:total-loss}
\mathcal{L}_{\mathrm{total}}(\theta,\phi)
= \mathcal{L}_{\pi}(\theta)
+ \beta_V\,\mathcal{L}_{V}(\phi)
+ \beta_H\,\mathcal{L}_{H}(\theta)
+ \beta_{\mathrm{spline}}\,\mathcal{R}_{\mathrm{spline}}.
\end{equation}
Advantage normalization, gradient clipping, and mini-batch epochs follow standard PPO practice. Reward shaping uses a compact, state-centric design balancing forward progress, energy efficiency, and safety (coefficients reported in the implementation section).

\subsection{Multi-Modal Delay Randomization (MMDR)}
Building on the observation in \eqref{eq:obs}, we inject modality-specific delays during training and then use the \emph{same} encoders/fusion pipeline on the delayed inputs \citep{imai2022vision}.

\paragraph{Proprioception latency.}
Let the simulator step be $\delta$. For each episode, sample $\Delta^{\mathrm{prop}}\!\sim\!\mathcal{U}[0,\Delta_{\max}]$ and set
\begin{align}
k      &= \Big\lfloor \Delta^{\mathrm{prop}}/\delta \Big\rfloor,\\
\alpha &= \Delta^{\mathrm{prop}}/\delta - k \in [0,1).
\end{align}
Using a FIFO buffer of past proprioceptive states,
\begin{equation}
\tilde s^{\mathrm{prop}}_t
=(1-\alpha)\, s^{\mathrm{prop}}_{t-k}\;+\;\alpha\, s^{\mathrm{prop}}_{t-k-1}.
\end{equation}

\paragraph{Visual latency.}
Maintain a buffer of the most recent $4k$ depth frames, partition it into four contiguous blocks of length $k$, and sample one index per block to compose a 4-frame stack:
\begin{equation}
\begin{aligned}
i_j &\sim \mathcal{U}\{(j-1)k,\ldots,jk-1\},\quad j=1,\ldots,4,\\
\tilde I^{\mathrm{depth}}_t
&= \big[I_{t-i_1},\, I_{t-i_2},\, I_{t-i_3},\, I_{t-i_4}\big].
\end{aligned}
\end{equation}

\paragraph{Policy input.}
We simply replace $o_t$ in \eqref{eq:obs} by the delayed input
\begin{equation}
\tilde o_t
=\big\{\,\tilde s^{\mathrm{prop}}_t,\ \tilde I^{\mathrm{depth}}_t\,\big\},
\end{equation}
and feed $\tilde o_t$ through the same encoders $f_{\mathrm{KAN}}(\cdot)$ and $f_{\mathrm{CNN}}(\cdot)$ as in \eqref{eq:prop-encoding}--\eqref{eq:vis-encoding}, followed by the unchanged projection, concatenation and the backbone.

\bigskip
\section{Implementation}
\label{sec:implementation}

\subsection{Environment Setup}
All experiments are conducted on a workstation equipped with an Intel Xeon Gold 6430 CPU (128 cores, 2.1\,GHz base clock) and an NVIDIA GeForce RTX~4090 GPU (24\,GB, CUDA-12.6). The operating system is Ubuntu~22.04. The physics simulation is run with PyBullet \citep{coumans2021pybullet} and all models are implemented in Python 3.8 with PyTorch~2.4.1.

The model is evaluated in three simulated environments that differ in terrain difficulty and obstacle dynamics:
\begin{itemize}
  \item \textbf{Thin Obstacle}: a flat plane seeded with numerous thin cuboid obstacles.
  \item \textbf{Static Obstacle with Rugged Terrain}: uneven, discontinuous ground (maximum height 5\,cm), requiring careful foothold placement with random size cuboid obstacles.
  \item \textbf{Dynamic Obstacle with Rugged Terrain}: thin obstacles that move with randomly sampled velocities/directions on uneven discontinuous ground (maximum height 5\,cm).
\end{itemize}

\begin{figure}[t]
\centering
\begin{subfigure}{\linewidth}
\centering
\includegraphics[width=0.49\linewidth]{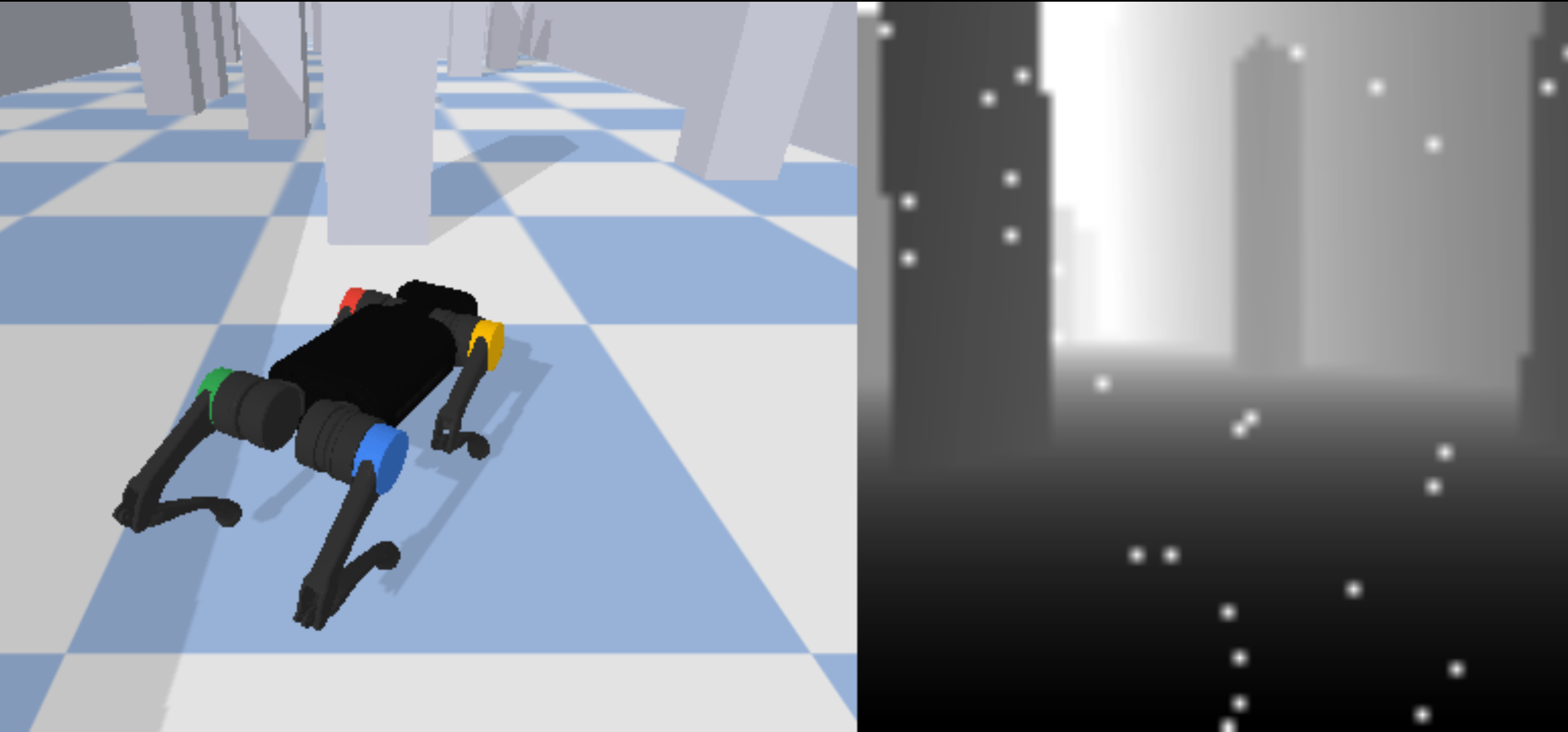}
\hfill
\includegraphics[width=0.49\linewidth]{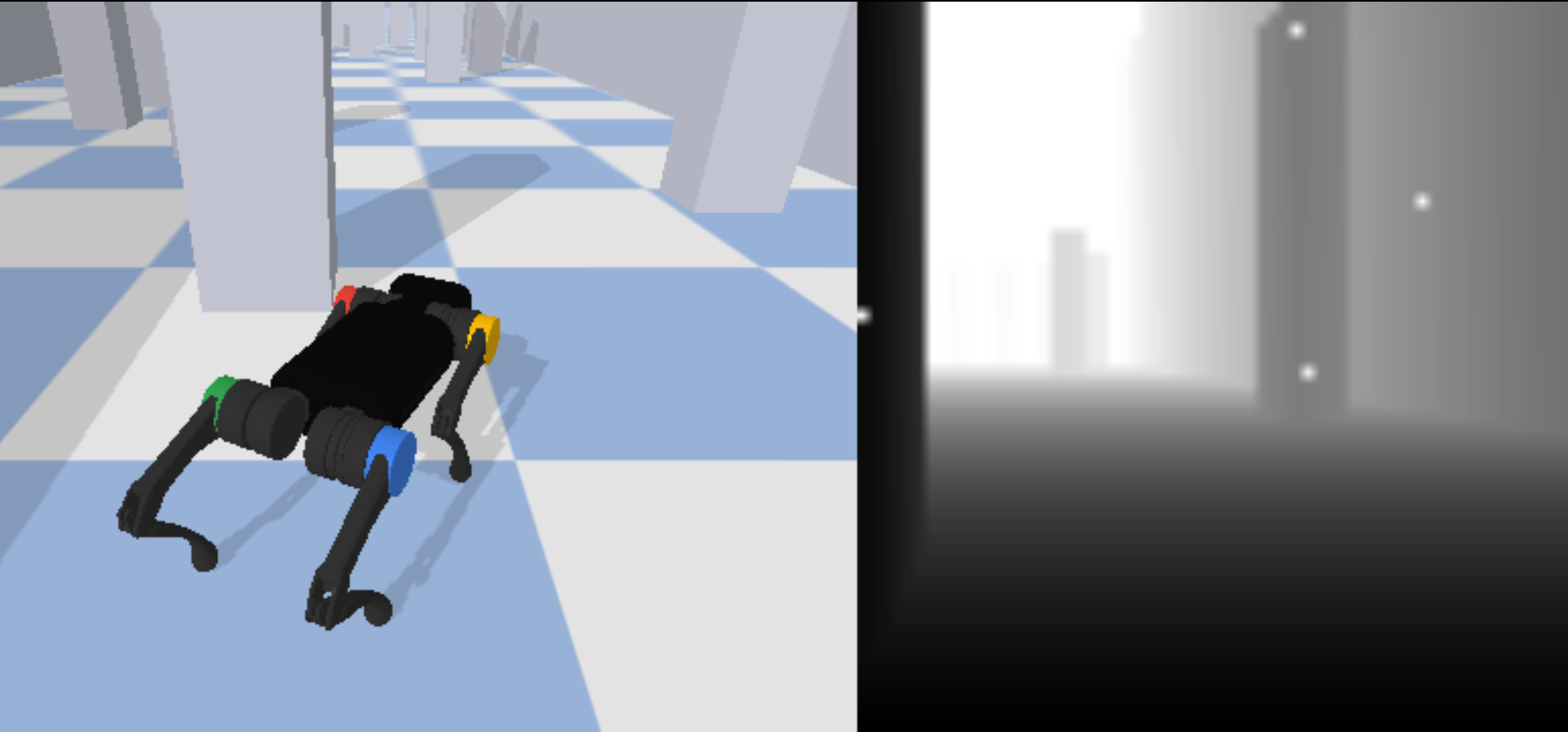}
    \caption{}
    \label{fig:suba}
\end{subfigure}
\begin{subfigure}{\linewidth}
\centering
\includegraphics[width=0.49\linewidth]{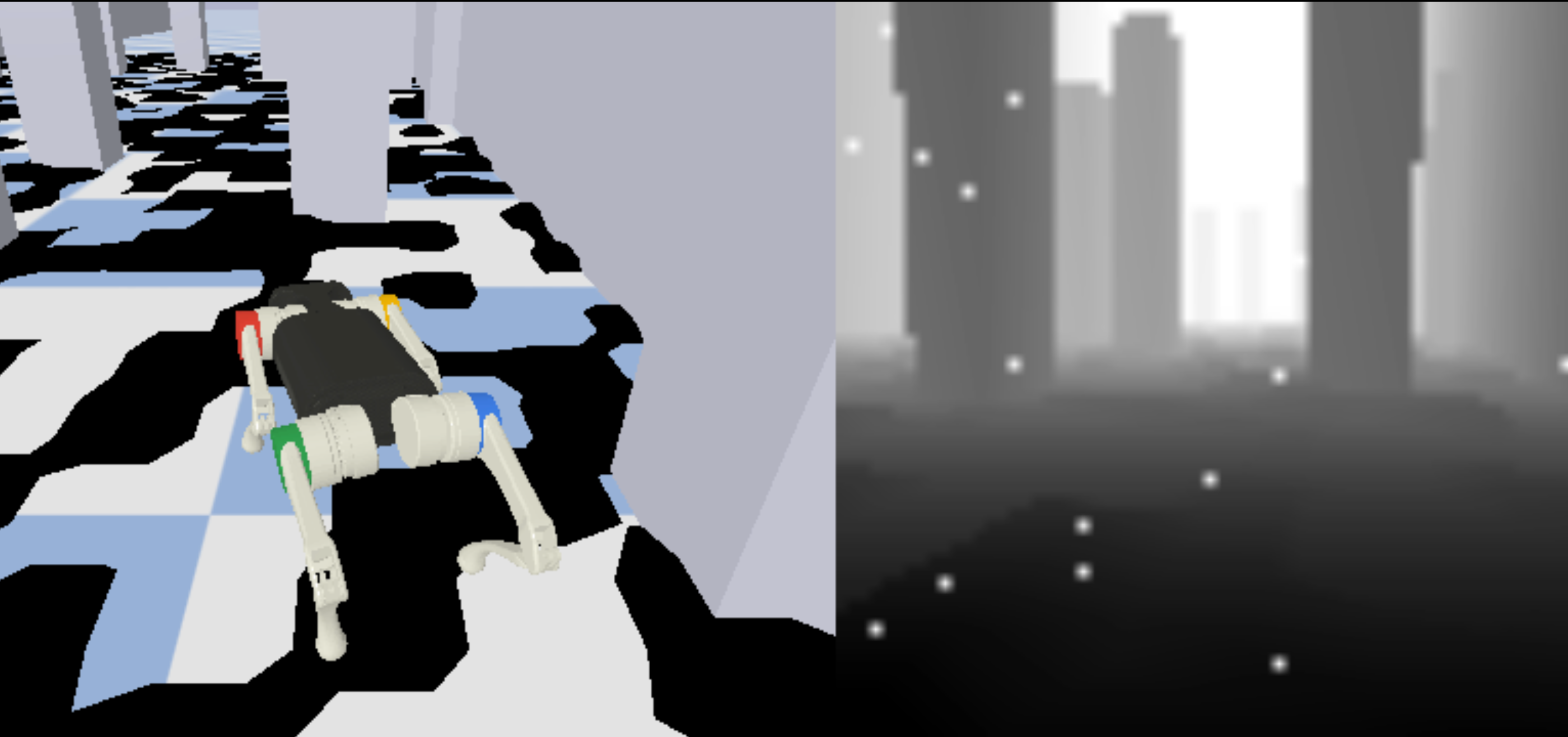}
\hfill
\includegraphics[width=0.49\linewidth]{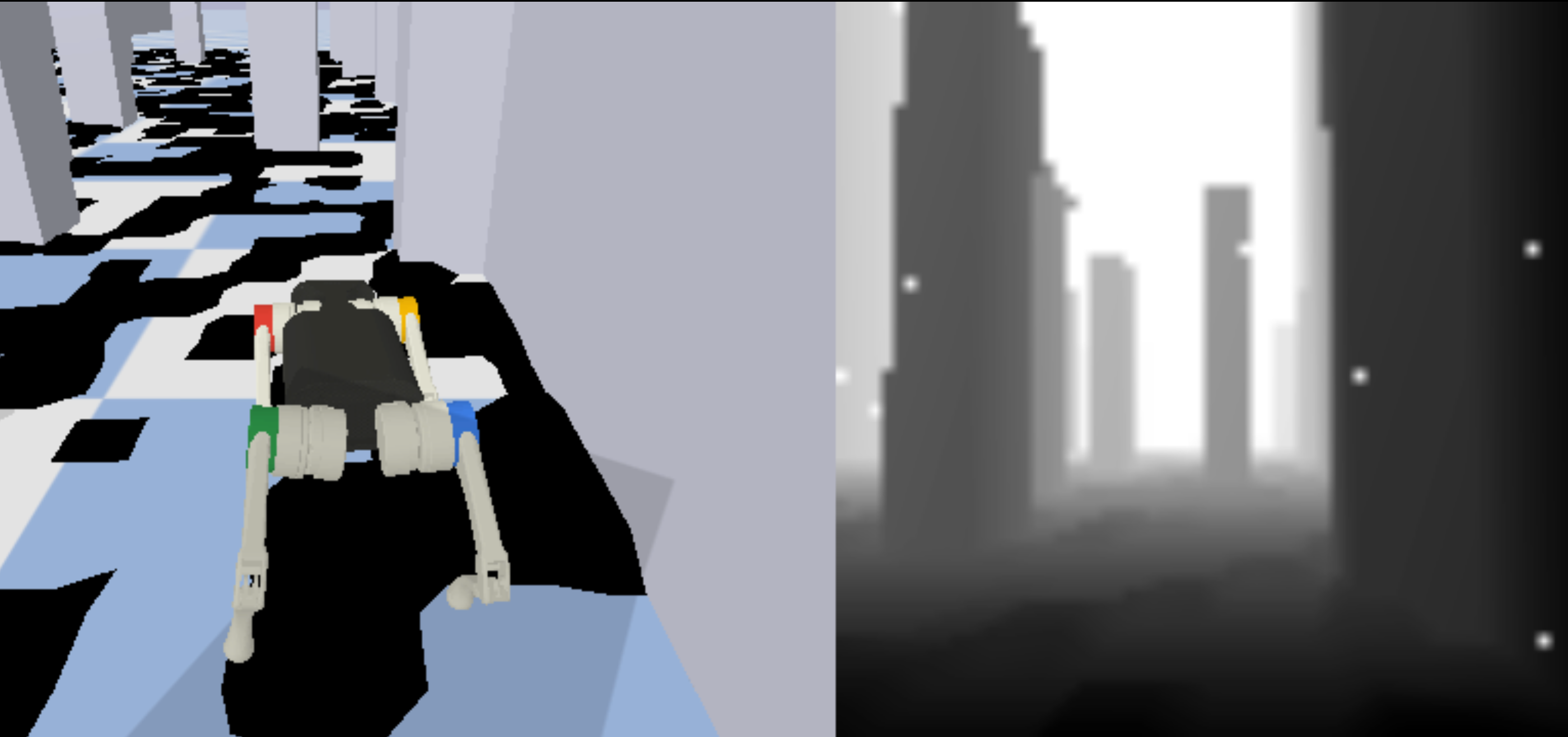}
    \caption{}
    \label{fig:subb}
\end{subfigure}
\begin{subfigure}{\linewidth}
\centering
\includegraphics[width=0.49\linewidth]{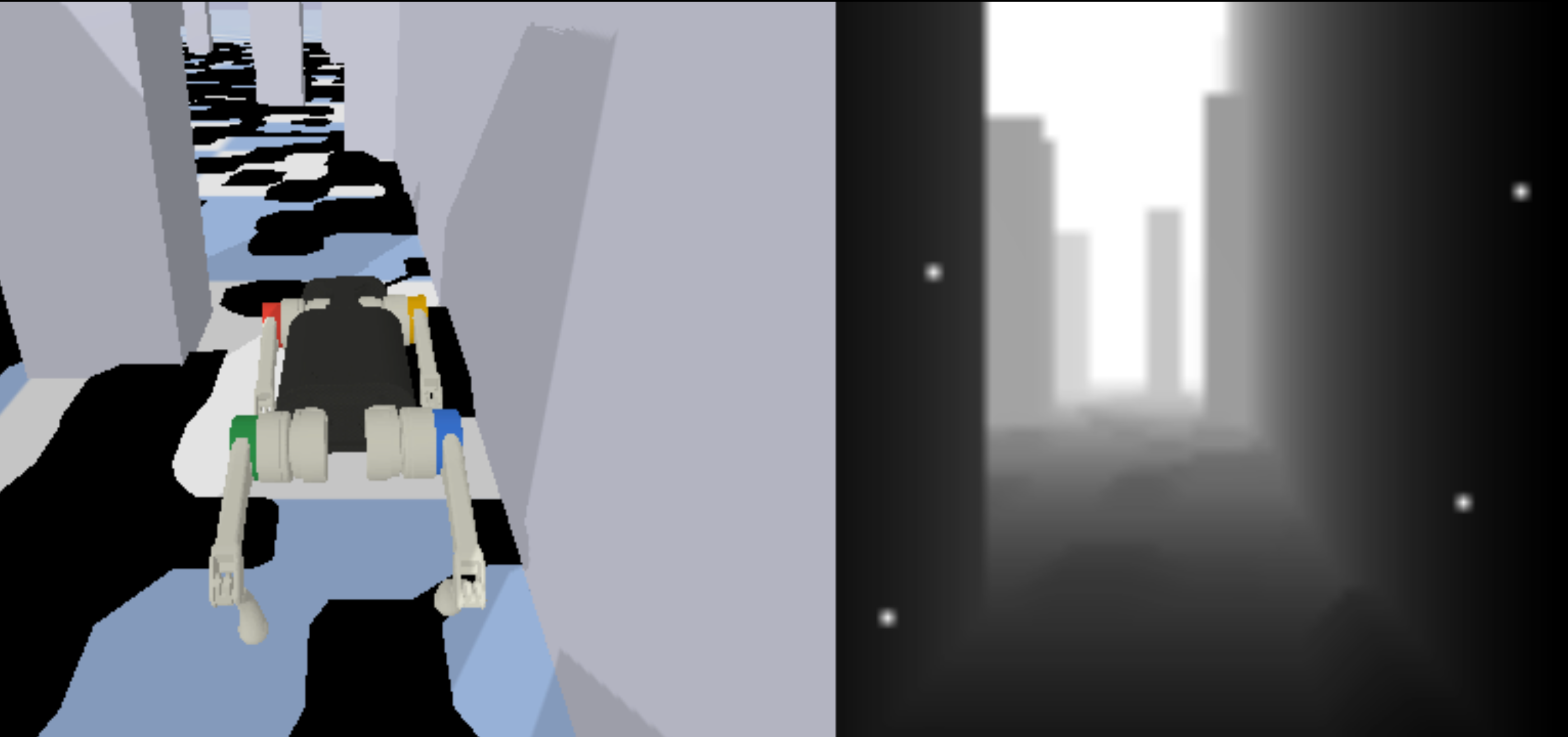}
\hfill
\includegraphics[width=0.49\linewidth]{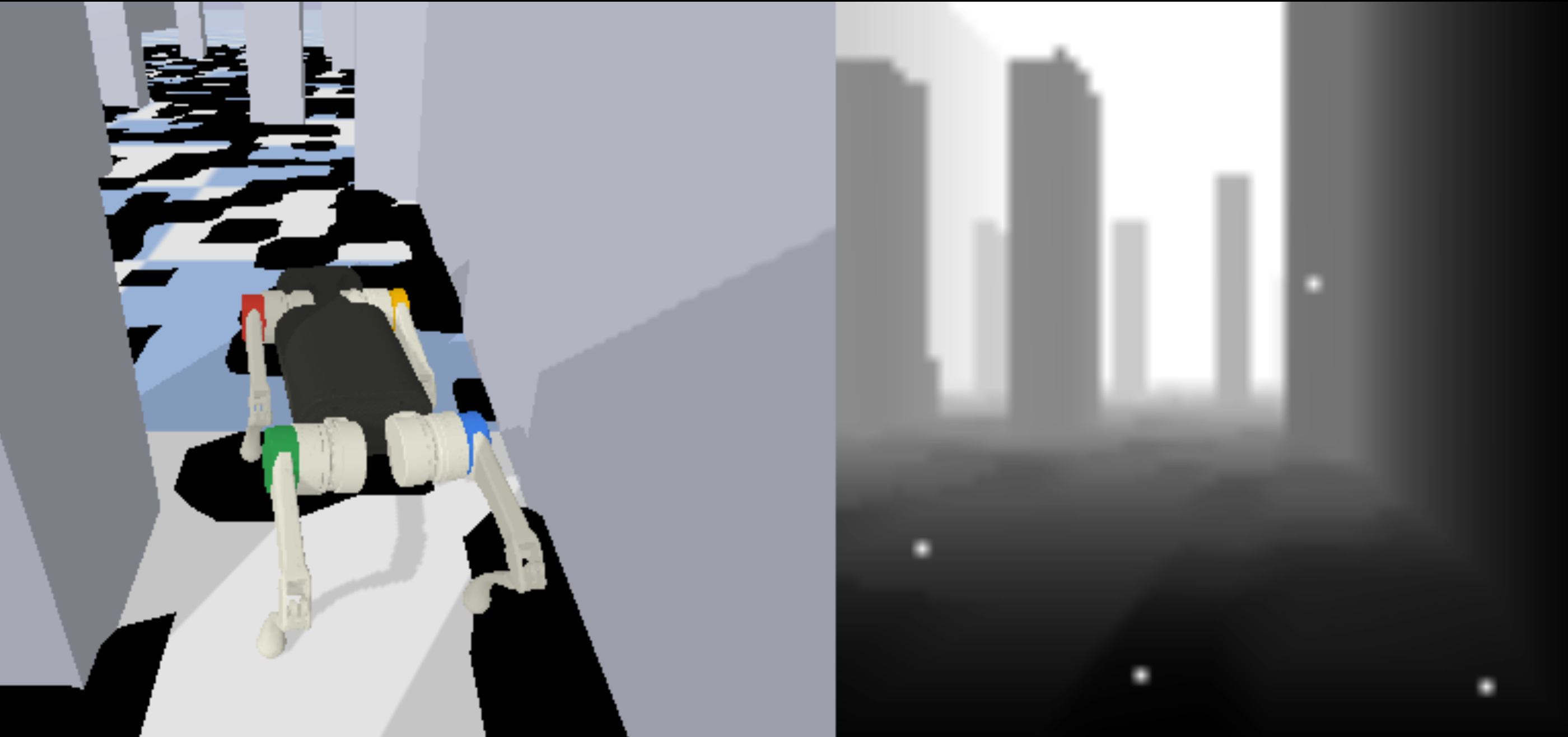}
    \caption{}
    \label{fig:subc}
\end{subfigure}
\caption{Evaluation environments: (a) \emph{Thin Obstacle}; (b) \emph{Static Obstacle with Rugged Terrain}; (c) \emph{Dynamic Obstacle with Rugged Terrain}. Layouts are randomized at reset; only (c) updates obstacle positions online during an episode.}
\label{fig:env_repre}
\end{figure}

Figure~\ref{fig:env_repre} shows representative scenes. Unless otherwise noted, obstacle layouts are resampled at episode reset; only \emph{Dynamic Obstacle with Rugged Terrain} updates obstacle positions during an episode.

\subsection{RL MDP Details}

\paragraph{Observation space.}
We adopt a standard multi-modal setting. Proprioception stacks the three most recent slices, yielding an $84$-D vector: per slice we include 12 joint angles, a 4-D IMU (roll/pitch and angular velocities), and the last executed 12-D action. Vision consists of four stacked $64{\times}64$ first-person depth images from a head-mounted sensor; depths are clipped to $[0.3,10]\,\mathrm{m}$.

\paragraph{Action space.}
Actions are 12-D desired joint position targets; a PD controller maps targets to torques at each control step.

\paragraph{Reward function.}
We use a compact, state-centric reward balancing forward progress, survival, and energy:
\begin{equation}
\label{eq:reward}
R_t \;=\; R^{\mathrm{fwd}}_t \;+\; 0.1\,R^{\mathrm{alive}}_t \;+\; 0.005\,R^{\mathrm{energy}}_t,
\end{equation}
with $R^{\mathrm{fwd}}_t=\langle v_t,e_x\rangle$, $R^{\mathrm{alive}}_t=1$ until termination, and $R^{\mathrm{energy}}_t=-\|\tau_t\|_2^2$, where $\tau_t$ are realized torques.

\subsection{Model Architecture Details}
Unless noted otherwise, widths follow the compact configuration in Table~\ref{tab:arch-quadkan}. The visual CNN is shared by policy and value heads. KAN modules use shallow depth and a small number of spline bases per unit to remain lightweight.

\begin{table}[t]
\centering
\small
\setlength{\tabcolsep}{6pt}
\caption{QuadKAN architecture settings.}
\label{tab:arch-quadkan}
\begin{tabularx}{\linewidth}{
  @{\hspace{4pt}} l
  >{\raggedright\arraybackslash}X
  @{\hspace{4pt}}
}
\toprule
\textbf{Component} & \textbf{Setting} \\
\midrule
Token width $d$ & 84 \\
Proprio encoder & 2-layer KAN; spline bases per unit $b{=}8$; hidden width 256 \\
Visual encoder & 3-layer CNN; patchify $64{\times}64$ with size $P{\times}P$ to $N{=}(64/P)^2$ tokens; project to $d$ \\
Fusion head & 3-layer KAN; input $(1{+}N)\!\times\! d$; pooled to a fused feature \\
Policy head & 3-layer KAN (256, 256) for $\mu_\theta$ and $\log\sigma_\theta$ \\
Value head & 3-layer KAN (256, 256) for $V_\phi$ \\
Regularization & Optional spline curvature and Jacobian/Lipschitz penalties on KAN coefficients \\
\bottomrule
\end{tabularx}
\end{table}

\subsection{Training Protocol}
We train with PPO using generalized advantage estimation, advantage normalization, gradient clipping, and early stopping based on a KL divergence threshold. Hyperparameters are shared across baselines (Table~\ref{tab:ppo-quadkan}). All methods train for $10\,\mathrm{M}$ environment steps with batch size $1{,}024$, split into 16 minibatches.

\begin{table}[t]
\centering
\caption{PPO hyperparameters.}
\label{tab:ppo-quadkan}
\begin{tabular*}{\linewidth}{
  @{\hspace{4pt}} l
  @{\extracolsep{\fill}} l
  @{\hspace{4pt}}
}
\toprule
\textbf{Hyperparameter} & \textbf{Value} \\
\midrule
Episode horizon (steps) & 999 \\
Samples per update & 16{,}384 \\
Minibatch size & 1{,}024 \\
Optimization epochs & 3 \\
Discount $\gamma$ & 0.99 \\
GAE $\lambda$ & 0.95 \\
PPO clip $\epsilon$ & 0.2 \\
Entropy coefficient & 0.005 \\
Policy/Value LR & $1\times 10^{-4}$ \\
Optimizer & Adam \\
Nonlinearity & ReLU \\
\bottomrule
\end{tabular*}
\end{table}

\paragraph{Obstacle-density curriculum.}
For obstacle scenarios, obstacle density is annealed linearly from an easy start to a target distribution over training iterations. The schedule is identical for all methods.

\subsection{Multi-Modal Delay Randomization (MMDR) Details}

\paragraph{Proprioception latency.}
Let $f_{\mathrm{sim}}$ and $f_{\mathrm{ctrl}}$ denote simulator and control rates, with steps $\delta_{\mathrm{sim}}{=}1/f_{\mathrm{sim}}$ and $\delta_{\mathrm{ctrl}}{=}1/f_{\mathrm{ctrl}}$ (e.g., $f_{\mathrm{sim}}{=}400$\,Hz, $f_{\mathrm{ctrl}}{=}25$\,Hz).
A FIFO buffer stores recent proprioceptive states at the simulator rate.
At episode start, sample a fixed delay $\Delta^{\mathrm{prop}}\!\sim\!\mathcal{U}[0,\Delta_{\max}]$ and decompose
\[
\Delta^{\mathrm{prop}}=k_{\mathrm{sim}}\delta_{\mathrm{sim}}+\alpha\,\delta_{\mathrm{sim}},
\quad k_{\mathrm{sim}}\in\mathbb{N}_0,\ \alpha\in[0,1).
\]
At a control step indexed by simulator time $t$, the delayed proprioception is the linear interpolation of adjacent buffer entries:
\begin{equation}
\tilde s^{\mathrm{prop}}_t
=(1-\alpha)\, s^{\mathrm{prop}}_{t-k_{\mathrm{sim}}}
\;+\;\alpha\, s^{\mathrm{prop}}_{t-k_{\mathrm{sim}}-1}.
\end{equation}
The sampled $\Delta^{\mathrm{prop}}$ remains fixed throughout the episode.

\paragraph{Visual latency.}
Let $f_{\mathrm{vis}}$ be the depth-camera frame rate (e.g., $30$\,Hz), with $\delta_{\mathrm{vis}}{=}1/f_{\mathrm{vis}}$.
Maintain a rolling buffer of the most recent $4k_{\mathrm{vis}}$ depth frames, split into four contiguous blocks of length $k_{\mathrm{vis}}$.

\subsection{Domain Randomization and Visual Perturbations}
At each episode reset, we sample physics and control parameters independently within the ranges in Table~\ref{tab:dr-quadkan}.
Separately from MMDR, we inject mild depth artifacts by drawing $K\!\sim\!\mathcal{U}\{3,\ldots,30\}$ pixels per frame and setting them to the sensor's maximum measurable depth (10\,m), mimicking stochastic dropouts.

\begin{table}[t]
\centering
\small
\setlength{\tabcolsep}{6pt}
\caption{Domain randomization ranges (episode-wise i.i.d. draws unless noted).}
\label{tab:dr-quadkan}
\begin{tabular*}{\linewidth}{
  @{\hspace{4pt}} l
  @{\extracolsep{\fill}} l
  l
  @{\hspace{4pt}}
}
\toprule
\textbf{Parameter} & \textbf{Range} & \textbf{Units} \\
\midrule
Body mass (per link)            & $[0.8,\ 1.2]\times$ default         & -- \\
Link inertia (per principal)    & $[0.5,\ 1.5]\times$ default         & -- \\
Friction coefficient $\mu$      & $[0.5,\ 1.25]$                      & -- \\
Motor viscous friction          & $[0.0,\ 0.05]$                      & N$\cdot$m$\cdot$s/rad \\
Motor strength (torque limit)   & $[0.8,\ 1.2]\times$ default         & -- \\
$K_P$ (joint PD)                & $[40,\ 90]$                         & -- \\
$K_D$ (joint PD)                & $[0.4,\ 0.8]$                       & -- \\
Proprioception latency          & $[0,\ 0.04]$                        & s \\
\bottomrule
\end{tabular*}
\end{table}

\section{Experimental Evaluation}
\label{sec:evaluation}

\subsection{Evaluation Setup}

\paragraph{Research Questions}
Our evaluation is organized around the following research questions (RQs):
\begin{itemize}
  \item \textbf{RQ1 (Cross-modal necessity).} Does combining proprioception and depth under the same delay model outperform either single modality on return, collisions, and distance across both seen and unseen terrains?
  \item \textbf{RQ2 (KAN effectiveness).} At matched parameter count and runtime, do KAN-based proprioceptive and/or fusion heads improve sample efficiency and performance, while reducing action jitter and energy, compared with unstructured regressors?
  \item \textbf{RQ3 (Training robustness).} Does the proposed end-to-end PPO training scheme improve learning stability and performance consistency?
\end{itemize}

\paragraph{Evaluation Metrics}
We report (i) \textit{mean episode return}, and two locomotion-specific measures: (ii) \textit{distance moved}, the net displacement (m) along the task-aligned axis, and (iii) \textit{collision times}, the number of robot–obstacle contacts accumulated during evaluation. Collisions are checked at every control step and aggregated until either three evaluation episodes complete or the robot falls. The collision metric is reported only in obstacle-containing scenarios and only for episodes that experience at least one obstacle interaction.

\paragraph{Baselines and Variants}
To assess the effectiveness of \emph{QuadKAN}, we compare against the following configurations. All agents share the same PPO/GAE setup, curriculum, domain randomization, and the same proprioceptive and depth encoders with a matched token width to minimize representation-size confounds.

\begin{itemize}
  \item \textbf{Proprio-Only.} Uses proprioception alone with an MLP policy/value head; no exteroceptive input.
  \item \textbf{MLP Vision-Only.} Uses depth alone with the shared CNN encoder; an MLP head consumes the visual features.
  \item \textbf{KAN Vision-Only.} Uses depth alone with the shared CNN encoder; a spline-parameterized KAN head (no proprioception).
  \item \textbf{MLP Proprio–Vision.} Encodes proprioception and depth with the same encoders as ours, projects both to the same width, concatenates the vectors, and feeds them to an \emph{unstructured} MLP fusion/policy head.
  \item \textbf{QuadKAN (ours).} KAN-based proprioceptive encoder and KAN-based fusion head (attention-free, spline-parameterized cross-modal policy).
\end{itemize}

\subsection{Performance Results on Simulation Scenarios}
\begin{figure}[t]
  \centering
  \includegraphics[width=\linewidth]{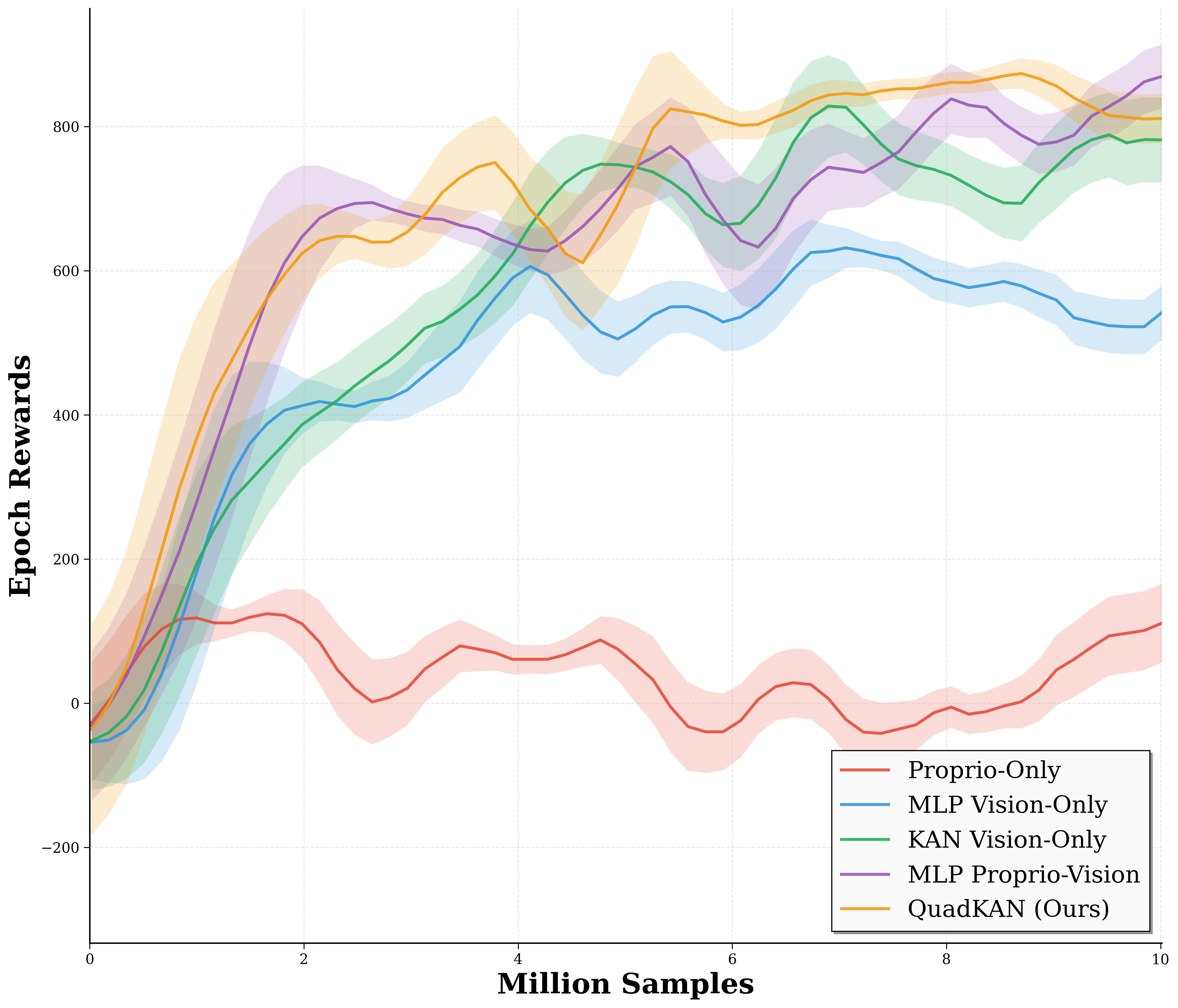}
  \caption{Training learning curves on \emph{Thin Obstacle}. Solid lines denote the mean episode return across seeds; shaded regions indicate $\pm$ one standard deviation.}
  \label{fig:learn_curves}
\end{figure}

Fig. \ref{fig:learn_curves} shows training curves with mean and one standard deviation across seeds. Table~\ref{tab:performance_comparison} reports means and standard deviations over seeds for episode return, collisions per episode, and progress distance. 
The supplementary video is available at \url{https://github.com/allen-quad-robot/quadkan}.

From Fig.~\ref{fig:learn_curves}, \emph{QuadKAN} converges faster (steeper early-learning slope), attains a higher asymptotic return, and exhibits lower inter-seed variance than all baselines. Relative to the best unstructured fusion baseline (\textit{MLP Proprio–Vision}), \emph{QuadKAN} improves return by \textbf{10.1\%} (874.37 vs.\ 794.30), reduces collisions by \textbf{75.9\%} (15.23 vs.\ 63.27), and attains a comparable distance (+\textbf{1.3\%}; 37.50 vs.\ 37.03), see Table~\ref{tab:performance_comparison}. Compared with \textit{Proprio-Only}, QuadKan got \textbf{+632.1\%} return, \textbf{--96.3\%} collisions, and \textbf{+674.8\%} distance. Vision-only policies traverse very short distances on average; collision counts are therefore not informative and are omitted. These trends (i) confirm the necessity of combining proprioception and depth (\textbf{RQ1}); (ii) demonstrate that KAN-based proprioceptive and/or fusion heads improve sample efficiency and performance while reducing action jitter and energy compared with unstructured regressors (\textbf{RQ2}); and (iii) indicate stable optimization under the proposed PPO training protocol (\textbf{RQ3}).

\begin{table*}[t]
\centering
\begin{minipage}{0.92\linewidth}
  \centering
  \caption{Performance on trained terrain (mean $\pm$ std over 10 runs).}
  \label{tab:performance_comparison}
  \small
  \begin{tabular*}{\linewidth}{
    @{\hspace{4pt}} l
    @{\extracolsep{\fill}} c c c
    @{\hspace{4pt}}
  }
    \toprule
    \textbf{Method} & \textbf{Return} & \textbf{Collisions} & \textbf{Distance (m)} \\
    \midrule
    Proprio-Only         & $119.43 \pm 37.26$  & $408.67 \pm 160.79$ & $4.84 \pm 1.09$ \\
    MLP Vision-Only      & $12.55 \pm 44.37$   & --                  & $1.27 \pm 1.33$ \\
    KAN Vision-Only      & $-11.40 \pm 10.96$  & --                  & $0.58 \pm 0.32$ \\
    MLP Proprio-Vision   & $794.30 \pm 98.71$  & $63.27 \pm 86.91$   & $37.03 \pm 5.27$ \\
    \textbf{QuadKAN (Ours)} & $\mathbf{874.37 \pm 34.06}$ & $\mathbf{15.23 \pm 20.69}$ & $\mathbf{37.50 \pm 2.28}$ \\
    \bottomrule
  \end{tabular*}
\end{minipage}
\end{table*}

\subsection{Albation Studies}
\label{sec:ablation}

\paragraph{Ablation on modalities.}
With identical encoders and PPO settings, \emph{MLP Proprio–Vision} already outperforms \emph{Proprio-Only} by \textbf{+565.1\%} in return (794.30 vs.\ 119.43), reduces collisions by \textbf{--84.5\%} (63.27 vs.\ 408.67), and extends distance by \textbf{+665.1\%} (37.03\,m vs.\ 4.84\,m). 
This substantial improvement confirms that proprioception and vision provide complementary cues—egocentric stability from proprioception and exteroceptive awareness from depth—making their fusion essential for obstacle negotiation.  

Building on this fusion baseline, \emph{QuadKAN} further improves performance to $874.37$ return, $15.23$ collisions, and $37.50$\,m distance, yielding \textbf{+10.1\%} higher return and \textbf{--75.9\%} fewer collisions than \emph{MLP Proprio–Vision}. 
These gains indicate that the spline-parameterized function class not only leverages cross-modal input more effectively but also enhances safety and stability.  
These results answer \textbf{RQ1} (fusion necessity) and reinforce \textbf{RQ2} (effectiveness of spline-structured policies).

\paragraph{Ablation on the fusion function class (KAN vs.\ MLP).}
Table~\ref{tab:performance_comparison} contrasts spline-structured fusion against unstructured MLPs. 
In the proprioception–vision setting, \emph{QuadKAN} achieves the highest performance across all metrics, improving over \emph{MLP Proprio–Vision} by \textbf{+10.1\%} in return (874.37 vs.\ 794.30), reducing collisions by \textbf{--75.9\%} (15.23 vs.\ 63.27), and slightly increasing progress distance (\textbf{+1.3\%}). 
This shows that introducing spline structure in both the proprioceptive encoder and fusion head yields not only higher reward but also substantially safer behavior. These results support \textbf{RQ2}: the spline-structured policy improves effectiveness over unstructured regressors under identical configuration/training conditions.

\subsection{Training Stability Analysis}
\label{sec:stability}
\begin{figure}[t]
\centering
\includegraphics[width=\linewidth]{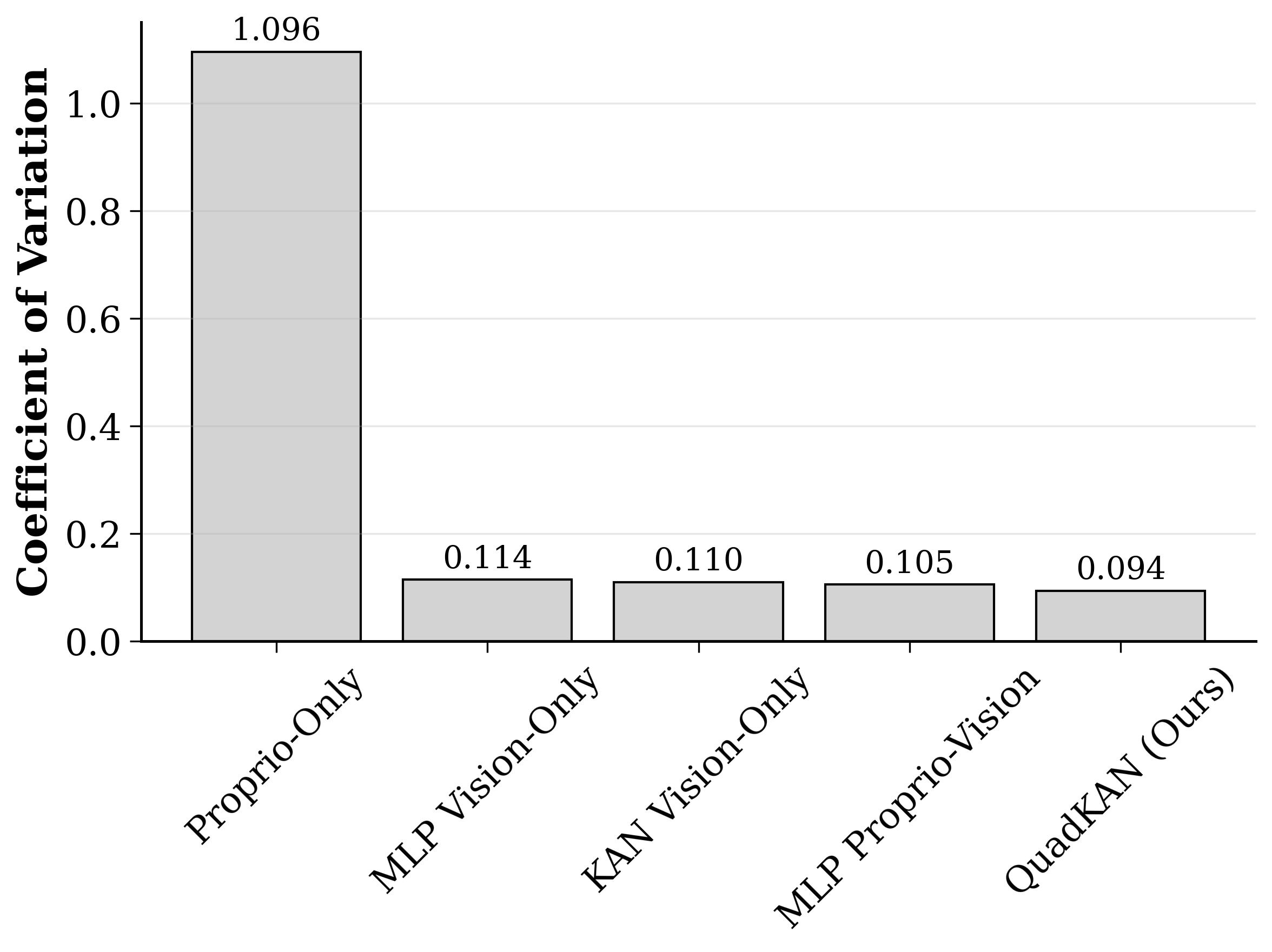} 
\caption{Training stability measured by the coefficient of variation (CoV) of all episodes return. Lower values indicate more stable optimization.}
\label{fig:training_stability}
\end{figure}

Fig.~\ref{fig:training_stability} reports the coefficient of variation (CoV = std/mean) of the episode return over the whole training epochs. A lower CoV indicates more stable optimization across seeds.

\emph{QuadKAN} achieves the lowest CoV ($0.094$), indicating the most stable training dynamics. Relative to \emph{MLP Proprio–Vision}, CoV decreases from $0.105$ to $0.094$ (a $\sim\!10.5\%$ reduction). Both \emph{MLP Vision-Only} ($0.114$) and \emph{KAN Vision-Only} ($0.110$) exhibit moderate stability but very limited locomotion performance, making their stability less meaningful. In contrast, \emph{Proprio-Only} suffers from a much higher CoV ($1.096$), reflecting highly volatile optimization due to partial observability and lack of foresight.

These results confirm that the spline-parameterized fusion in \emph{QuadKAN} not only improves performance but also stabilizes training under PPO, supporting both the effectiveness of structured policies (\textbf{RQ2}) and the robustness of the overall training scheme (\textbf{RQ3}).

\subsection{Generalization to Unseen Conditions}
\label{sec:generalization}
The policies trained in the \emph{Thin Obstacle} environment are evaluated zero-shot on two previously unseen scenarios: a static random-shape obstacle terrain and a moving-obstacle terrain with rugged ground. Table~\ref{tab:challenging_terrain_comparison} reports mean and standard deviation over 10 runs. For vision-only baselines, collision counts are omitted because these agents cover little distance and collisions are not informative.

\paragraph{Static Obstacles with Rugged Terrain.}
Cross-modal policies generalize substantially better than single-modality baselines. \emph{QuadKAN} achieves the best overall performance, with the highest return ($787.98 \pm 191.89$), the fewest collisions ($35.17 \pm 24.47$), and near-maximal distance ($36.13 \pm 4.74$). Relative to \emph{Proprio-Only}, it improves return more than sixfold, reduces collisions by over $93\%$, and increases distance by nearly $591\%$. Compared with \emph{MLP Proprio–Vision}, \emph{QuadKAN} further yields a \textbf{+5.7\%} higher return, \textbf{--44.4\%} fewer collisions, and a comparable distance (\textbf{--1.2\%}). Vision-only models fail to generalize, producing negative or near-zero rewards and minimal displacement. These results confirm that proprioception–vision fusion is necessary for robust out-of-distribution generalization (\textbf{RQ1}), and that spline-parameterized fusion further improves both safety and stability under unseen obstacles (\textbf{RQ2}).

\paragraph{Moving Obstacles with Rugged Terrain.}
Generalization becomes more challenging under dynamic hazards and uneven ground. \emph{QuadKAN} again outperforms all baselines, achieving the highest return ($261.52 \pm 371.47$), the fewest collisions ($395.03 \pm 96.80$), and the greatest progress distance ($17.77 \pm 8.07$). Compared with \emph{Proprio-Only}, this corresponds to a $+78\%$ increase in return, a $--8.4\%$ reduction in collisions, and a $+220.7\%$ increase in distance. Relative to \emph{MLP Proprio–Vision}, \emph{QuadKAN} delivers a dramatic $+147.4\%$ gain in return, reduces collisions by $--8.7\%$, and achieves a $+41.0\%$ longer distance (Table~\ref{tab:challenging_terrain_comparison}). By contrast, vision-only variants collapse completely, producing near-zero or negative rewards with negligible movement. These findings highlight that KAN-enhanced proprioception–vision fusion within PPO is critical for negotiating dynamic, partially observable environments, and that the structured spline formulation provides tangible advantages in both robustness and stability (\textbf{RQ2}, \textbf{RQ3}).

\begin{table*}[t]
\centering
\begin{minipage}{\linewidth}
  \centering
  \caption{Performance comparison of different model configurations on unseen terrains (mean $\pm$ std over 10 runs).}
  \label{tab:challenging_terrain_comparison}
  \small
  \begin{tabular*}{\linewidth}{
    @{\hspace{4pt}} l
    @{\extracolsep{\fill}} c c c  c c c
    @{\hspace{4pt}}
  }
    \toprule
    & \multicolumn{3}{c|}{\textbf{Static Obstacle with Rugged Terrain}} & \multicolumn{3}{c}{\textbf{Dynamic Obstacle with Rugged Terrain}} \\
    \cmidrule(lr){2-4} \cmidrule(lr){5-7}
    \textbf{Models} \\ & \textbf{Reward} & \textbf{Collision} & \textbf{Distance (m)} & \textbf{Reward} & \textbf{Collision} & \textbf{Distance (m)} \\
    \midrule
    Proprio-Only         & $122.46 \pm 45.02$   & $510.83 \pm 126.38$ & $5.22 \pm 1.65$  & $146.68 \pm 66.94$   & $431.20 \pm 207.53$ & $5.54 \pm 2.18$ \\
    MLP Vision-Only      & $15.93 \pm 39.51$    & --                  & $1.27 \pm 1.21$  & $-5.80 \pm 32.21$    & --                  & $0.68 \pm 1.09$ \\
    KAN Vision-Only      & $-11.38 \pm 11.08$   & --                  & $0.54 \pm 0.34$  & $-12.35 \pm 12.37$   & --                  & $0.66 \pm 0.44$ \\
    MLP Proprio-Vision   & $745.26 \pm 205.93$  & $63.20 \pm 50.33$   & $\mathbf{36.58 \pm 6.77}$ & $105.70 \pm 475.62$ & $432.53 \pm 115.01$ & $12.60 \pm 9.85$ \\
    \textbf{QuadKAN (Ours)} & $\mathbf{787.98 \pm 191.89}$ & $\mathbf{35.17 \pm 24.47}$ & $36.13 \pm 4.74$ & $\mathbf{261.52 \pm 371.47}$ & $\mathbf{395.03 \pm 96.80}$ & $\mathbf{17.77 \pm 8.07}$ \\
    \bottomrule
  \end{tabular*}
\end{minipage}
\end{table*}

\subsection{Spline Analysis and Interpretability}
\label{sec:spline_analysis}
Beyond aggregate performance metrics, we investigate the internal behavior of KAN-based policies to understand how spline parameterization contributes to effectiveness and stability \citep{ji2024comprehensive}. We analyze (i) weight-level patterns, (ii) activation-level nonlinearities, and (iii) comparative spline statistics across model configurations.

\begin{figure*}[t]
  \centering
  \includegraphics[width=\linewidth]{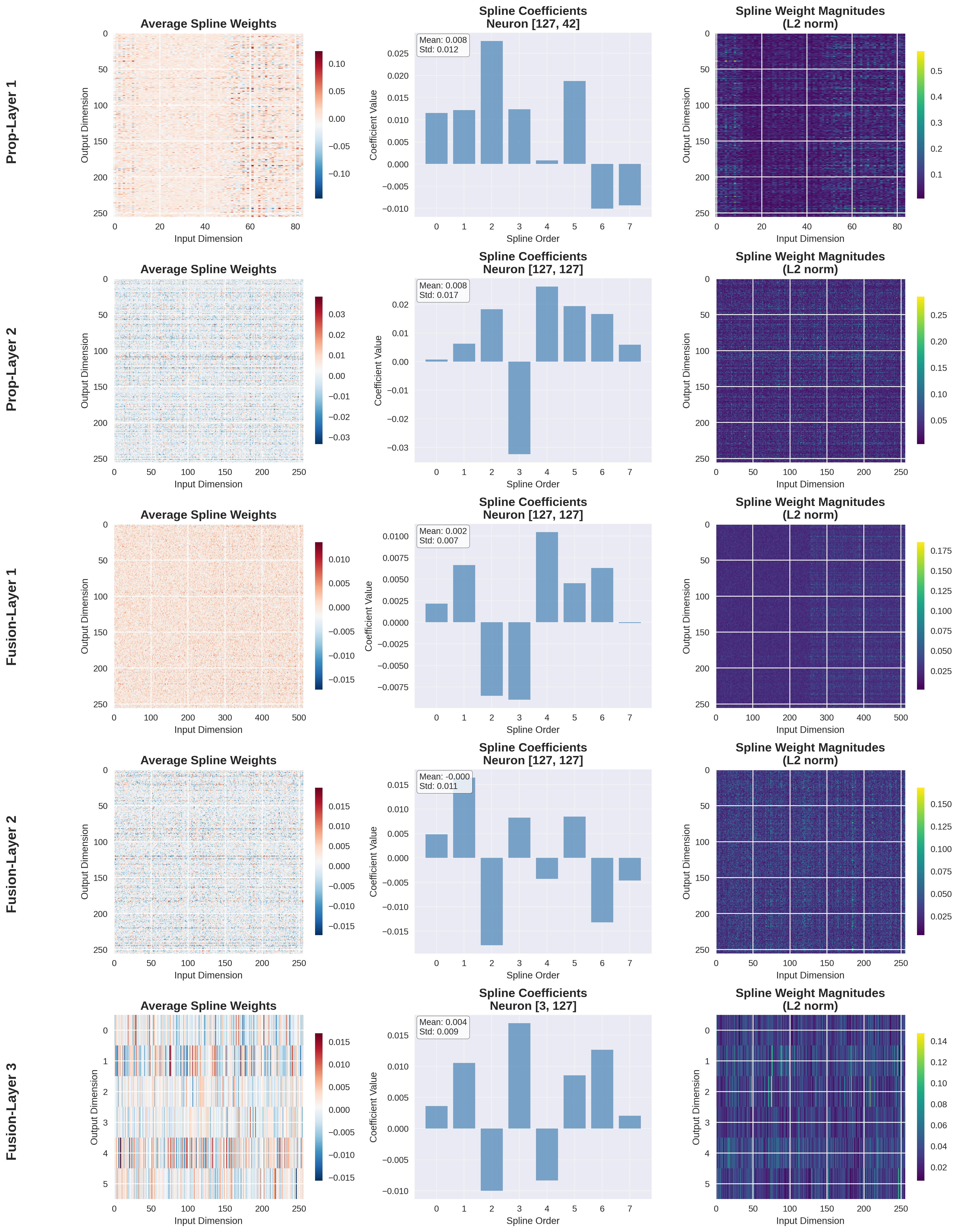}
  \caption{Spline weight analysis of \emph{QuadKAN (Ours)} (Including 2-layer Proprio encoder KAN and 3-layer Fusion head KAN as introduced in Table \ref{tab:arch-quadkan}). 
  Left: average spline weights across dimensions; 
  Middle: spline coefficient distributions for representative neurons; 
  Right: L2 norms of spline weight magnitudes with descriptive statistics. 
  Structured, localized weight patterns emerge, particularly in the proprioception–vision setting.}
  \label{fig:vision_state_kan_weights}
\end{figure*}

\begin{figure*}[h]
  \centering
  \includegraphics[width=\linewidth]{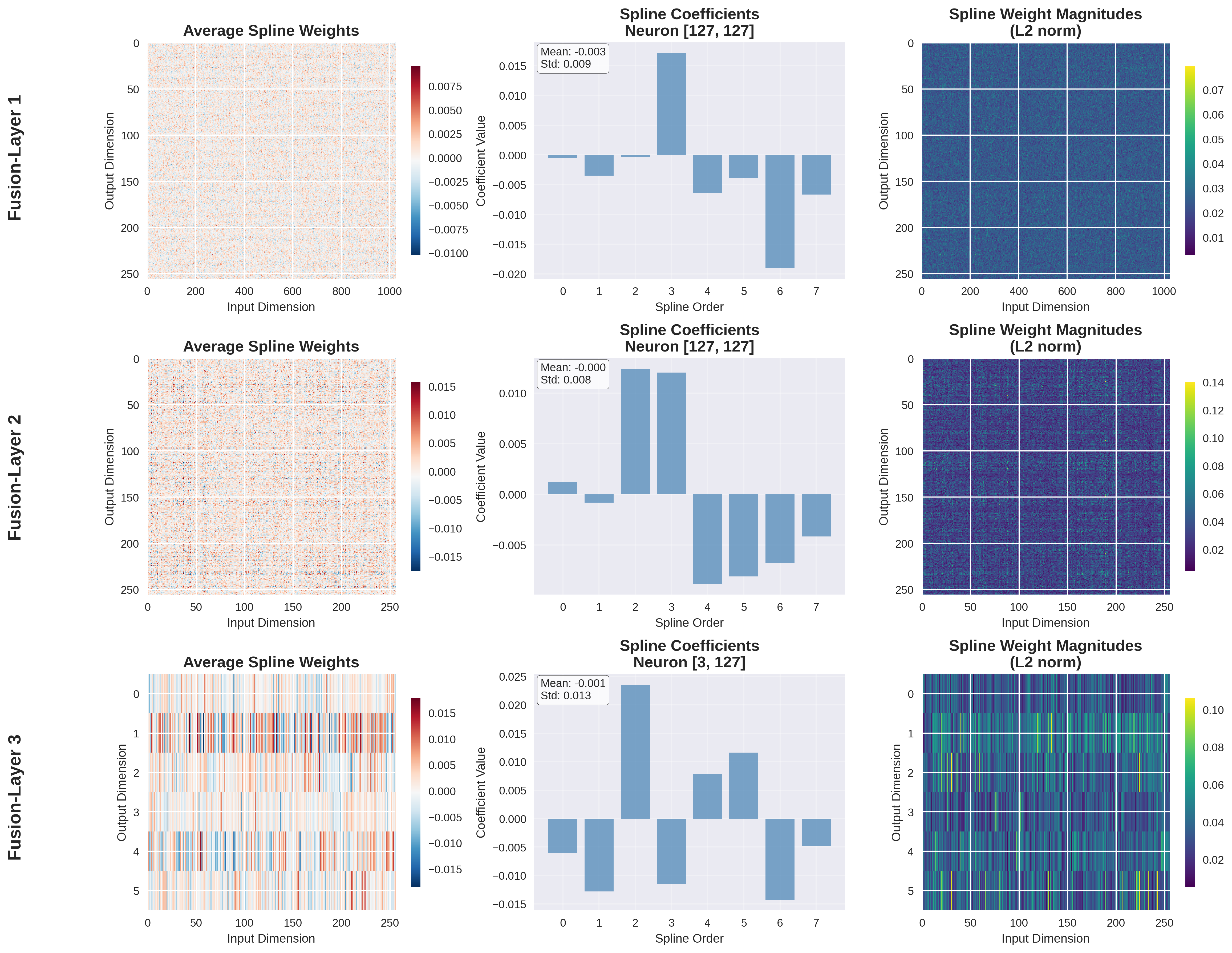}
  \caption{Spline weight analysis of \emph{Vision-Only KAN}. 
  Without proprioceptive anchoring, learned spline weights are weaker and less structured. 
  Compared with Fig.~\ref{fig:vision_state_kan_weights}, the representations show reduced localization and stability.}
  \label{fig:vision_only_kan_weights}
\end{figure*}

\paragraph{Weight-level analysis.}
Figures~\ref{fig:vision_state_kan_weights}--\ref{fig:vision_only_kan_weights} visualize spline weight distributions. 
In \emph{Vision-only KAN}, weights are structured and localized rather than dense and uniform, indicating that bases activate only in specific input regions. Compared with \emph{Vision-only KAN}, the \emph{QuadKAN}, utilizing proprioception–vision fusion, exhibits more pronounced and concentrated magnitudes, suggesting that proprioceptive anchoring stabilizes the learned bases. Coefficient histograms reveal balanced positive and negative responses, producing interpretable localized sensitivities aligned with phase-wise dynamics rather than opaque global couplings.

\begin{figure*}[t]
  \centering
  \includegraphics[width=\linewidth]{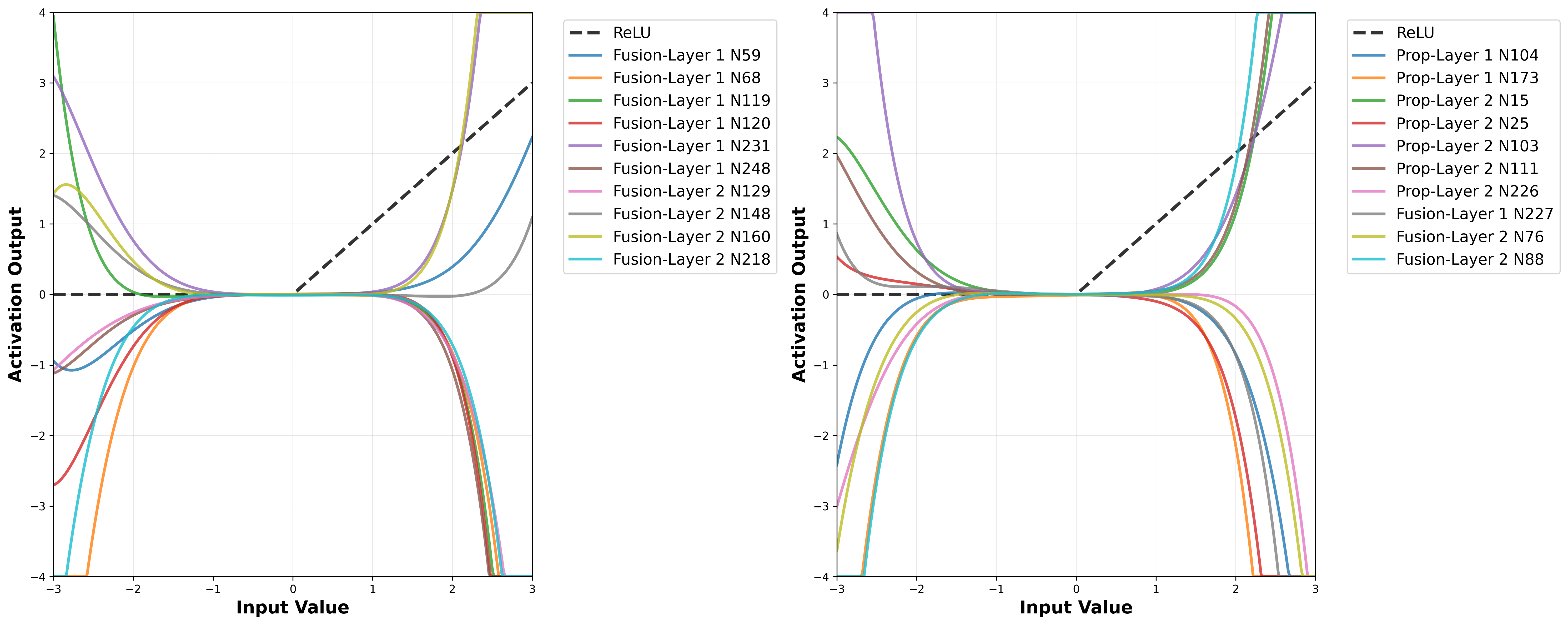}
  \caption{Spline activation functions (10 sampled neurons) vs.\ ReLU. Left: \emph{Vision-only KAN}; Right: \emph{QuadKAN (Ours)}. Unlike ReLU, spline activations exhibit rich nonlinearities to modal complex scenarios. Proprioceptive anchoring stabilizes activations, producing more coherent responses across neurons.}
  \label{fig:spline_activation_comparison}
\end{figure*}

\paragraph{Activation-level analysis.}
Figure~\ref{fig:spline_activation_comparison} compares spline activations against ReLU. While ReLU is piecewise-linear, spline activations exhibit richer nonlinearities with smooth curvature and local variations. In the \emph{Vision-only KAN}, neuron activations are more dispersed and heterogeneous, reflecting instability without proprioceptive anchoring. In contrast, the \emph{Vision–State KAN} produces more coherent and consistent activation patterns across neurons, indicating that proprioceptive input provides a stabilizing prior that regularizes spline responses.

\begin{figure*}[t]
  \centering
  \includegraphics[width=\linewidth]{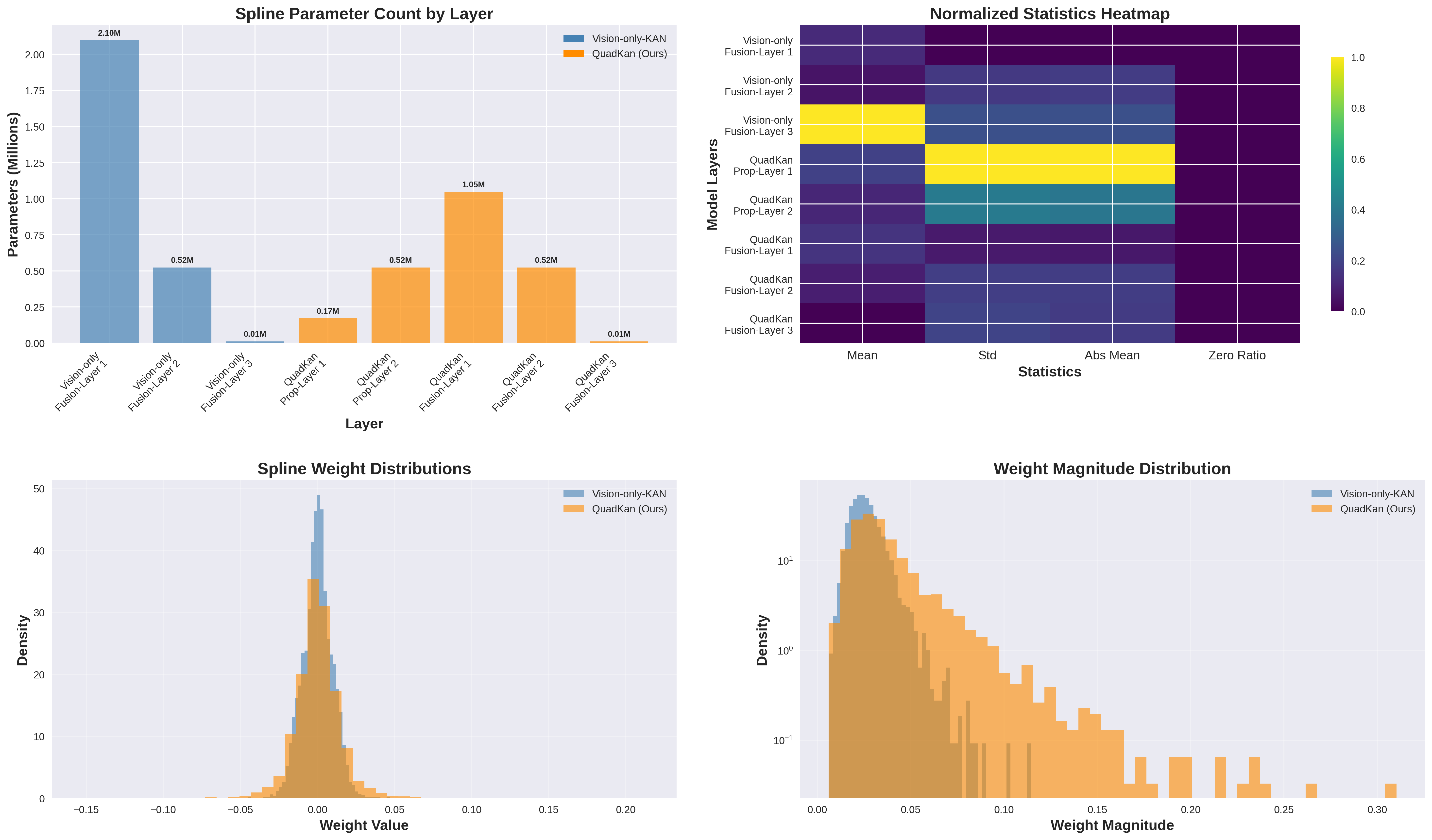}
  \caption{Comparative spline statistics for \emph{Vision-only KAN} and \emph{QuadKAN (Ours)}. 
  The dashboard summarizes parameter counts, weight distributions, normalized statistics, layer-wise magnitudes, and sample activations. 
  Despite fewer parameters, \emph{QuadKAN} learns richer and more coherent spline structures, aligning with its stability and performance gains.}
  \label{fig:kan_comparative}
\end{figure*}

\paragraph{Comparative spline statistics.}
Figure~\ref{fig:kan_comparative} summarizes global spline characteristics. 
Although \emph{Vision-only KAN} contains more parameters (2.63M vs.\ 2.22M), it does not generalize better, confirming that parameter count alone is insufficient. 
QuadKAN exhibits heavier-tailed weight distributions, larger absolute means, and greater variance across layers, indicating richer localized responses. 
The more uniform and widespread spline weight distribution further reveals more coherent behavior of QuadKAN, consistent with its observed performance and stability advantages.

Together, these analyses show that spline parameterization yields structured and interpretable representations, capturing complex nonlinear responses while maintaining local support. 
This interpretability complements the quantitative evidence from Section~\ref{sec:ablation}, supporting \textbf{RQ2} (structured KAN heads enhance performance and stability) and partially \textbf{RQ3} (training robustness is tied to more coherent spline parameterizations).

\section{Conclusion}
\label{sec:conclusion}

We presented \emph{QuadKAN}, a vision-guided DRL framework for quadrupedal locomotion that fuses proprioception with depth sensing and employs a spline-parameterized policy class. By aligning state–action mapping with the piecewise-smooth structure of gait, QuadKAN provides an attention-free fusion mechanism with favorable computational efficiency and exposes an interpretable structure through spline coefficients and localized sensitivities. Training with PPO under \emph{Multi-Modal Delay Randomization} (MMDR) further mitigates modality-dependent latencies and asynchrony.

Experiments addressed three research questions: \textbf{RQ1 (Cross-modal necessity)}—proprioception–vision fusion consistently outperforms unimodal baselines across return, distance, and collisions; \textbf{RQ2 (KAN effectiveness)}—spline-structured policies achieve faster convergence, lower jitter and energy proxies, and higher final performance than unstructured MLP fusion; and \textbf{RQ3 (Training robustness)}—the proposed training scheme improves stability and consistency. Together, these findings demonstrate that structured policies and training improve efficiency, robustness, and interpretability in vision-guided legged locomotion.

While this work focused on simulation, future efforts will deploy \emph{QuadKAN} on physical quadrupeds to assess sim-to-real transfer, latency handling, and safety under real-world conditions.

\bibliographystyle{model1-num-names}

\bibliography{cas-refs}


\end{document}